\documentclass[10pt]{cai}

\begin{document}
% Editorial staff will replace the following values:
% 1. Conference Year
% 2. Issue number
% 3. Article DOI
\begin{center}

\title{Intra-Layer Recurrence in Transformers for Language Modeling}
\maketitle

\thispagestyle{empty}

% Add Authors and Affiliations in the camera ready
% for the double blind review, please leave this section as is 
% \begin{tabular}{cc}
% Anthony Nguyen\upstairs{\affilone}, Wenjun Lin\upstairs{\affilone,*}
% \\[0.25ex]
% {\small \upstairs{\affilone} Digital Healthcare Innovation Lab \\  Faculty of Computer Sciene and Technology \\   Algoma University}
% \end{tabular}

\begin{tabular}{cc}
Anthony Nguyen\textsuperscript{\affilone}, Wenjun Lin\textsuperscript{\affilone,*}
\\[0.25ex]
\textsuperscript{\affilone} Digital Innovation Lab \\ Faculty of Computer Science and Technology \\ Algoma University, ON, Canada
\end{tabular}
  
% Replace with corresponding author email address
\emails{
  \upstairs{*}randy.lin@algomau.ca 
}
\vspace*{0.2in}
\end{center}

\begin{abstract}

Transformer models have established new benchmarks in natural language processing; however, their increasing depth results in substantial growth in parameter counts. While existing recurrent transformer methods address this issue by reprocessing layers multiple times, they often apply recurrence indiscriminately across entire blocks of layers. In this work, we investigate Intra-Layer Recurrence (ILR), a more targeted approach that applies recurrence selectively to individual layers within a single forward pass. Our experiments show that allocating more iterations to earlier layers yields optimal results. These findings suggest that ILR offers a promising direction for optimizing recurrent structures in transformer architectures.

\end{abstract}

% add your keywords
\begin{keywords}{Keywords:}
Transformers, Language Modeling, Intra-Layer Recurrence
\end{keywords}

\section{Introduction}

Transformer-based language models have achieved state-of-the-art performance across various NLP tasks \cite{vaswani2017attention, brown2020languagemodelsfewshotlearners}, but their increasing computational and memory demands present challenges. Architectural modifications that enhance performance without increasing parameter count are worth exploring.

A promising technique is to apply recurrence in transformers \cite{dehghani2019universal, giannou2023looped, yang2024looped, fan2024looped, geiping2025scaling}. However, past works apply this mechanism to the entire transformer model, reusing all layers multiple times per step, effectively increasing depth by a factor of two or more. While effective, this approach lacks granularity, treating all layers equally. We investigate Intra-Layer Recurrence (ILR), where select layers are re-entered independently within a single forward pass and allows finer control over effective depth.

\begin{figure}[h]
    \centering
    \includegraphics[width=0.3\linewidth]{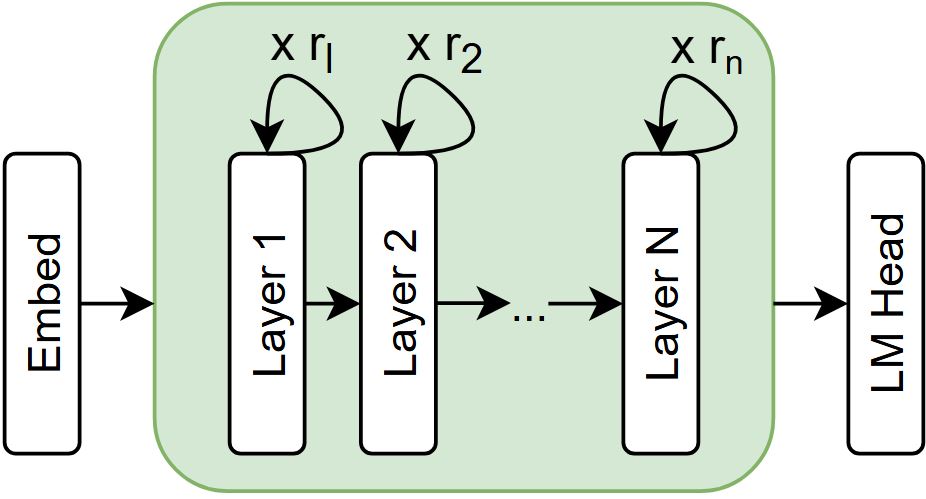}
    \caption{Transformer architecture with intra-layer recurrence.}
    \label{fig:1}
\end{figure}

This distinction is crucial, as different layers contribute uniquely to representations, and indiscriminate reuse of layers may not be optimal. By selectively reusing layers, we aim to determine which layers benefit the most. Furthermore, our experiments show that ILR still improves perplexity without increasing parameter count.

\section{Related Work}
\label{sec:related_work}

% Add tf-based lms (done)
% \subsection{Transformer-based Language Models}

The Transformer architecture \cite{vaswani2017attention} underpins modern language models, achieving state-of-the-art performance in NLP tasks. Unlike recurrent models such as LSTMs \cite{hochreiter1997long}, Transformers process tokens in parallel using self-attention and feedforward layers, entirely eliminating recurrence. This design improves scalability but comes at high computational and memory cost. Large-scale models like BERT \cite{devlin2019bert}, GPT \cite{radford2019language}, and LLaMA \cite{touvron2023llama} extend this approach, motivating research into more efficient architectures.

A potential approach to mitigating massive parameter scaling is the reintroduction of recurrence within transformers, a technique explored in previous works.
Universal Transformers \cite{dehghani2019universal} apply recurrence to a single-layer transformer, iterating multiple times to refine representations, in contrast to conventional transformers that process inputs through multiple distinct layers. Looped transformers \cite{giannou2023looped, yang2024looped, fan2024looped} extend this idea by applying recurrence to enhance algorithmic reasoning and length generalization. 

More recently, depth-recurrent transformers \cite{geiping2025scaling} structure recurrence into three blocks: a \textit{prelude}, a \textit{recurrent} block, and a \textit{coda}. The prelude and coda function as standard layer stacks, while the recurrent block iterates multiple times. This method applies recurrence to a block of layers, where all layers in the recurrent block iterate equally.

\begin{figure}[h]
    \centering
    \includegraphics[width=0.65\linewidth]{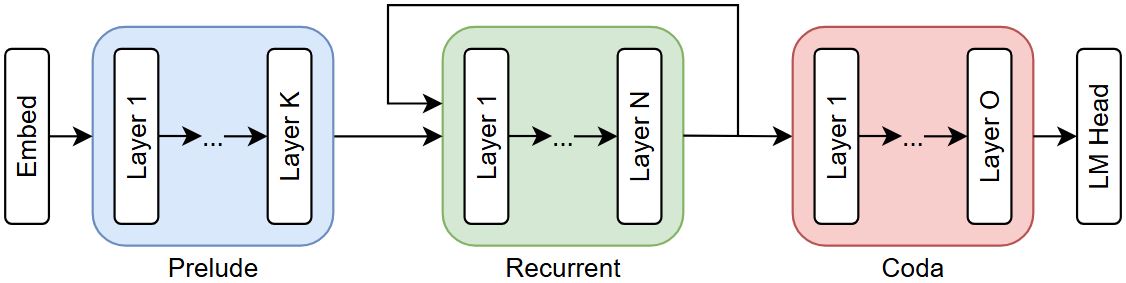}
    \caption{Depth-recurrent transformer proposed by Geiping et al.~\cite{geiping2025scaling}, which groups layers into three blocks and applies recurrence only to the middle block.}
    \label{fig:depth_recurrent_transformer}
\end{figure}

Unlike this approach, ILR applies recurrence at the individual layer level, selectively reusing layers within a single forward pass. This provides finer control over effective depth, allowing compute scaling without uniform recurrence across all layers. Our study investigates whether certain layers benefit more from recurrence, offering a more granular perspective on recurrent transformer efficiency.

Prior research suggests that transformer layers contribute differently to representation learning, with early layers capturing fundamental syntactic patterns and later layers introducing redundancy \cite{kovaleva2019bert}. 
Logit lens analysis \cite{logitlens2020} further reveals that token predictions become increasingly well-formed as they pass through early layers, suggesting that foundational representations emerge early and are progressively refined. 

These insights motivated our investigation into ILR, as they imply that if layers iteratively refine the representations produced by preceding layers, they may also benefit from recurrent self-refinement within a single layer.

\section{Methodology}
\label{sec:methodology}

A standard transformer with $L$ layers computes:
\[
h^{(l)} = f_\theta^{(l)}(h^{(l-1)}),
\]
where $h^{(l)}$ is the output of layer $l$, and $f_\theta^{(l)}$ is the transformation function parameterized by $\theta$.

Our approach introduces a \textit{reuse map} $\mathbf{R}=[r_1,\dots,r_L]$, where $r_l$ specifies the number of times layer $l$ is reused. The modified forward pass is:
\[
h^{(l,1)} = f_\theta^{(l)}(h^{(l-1)}), \quad h^{(l,k)} = f_\theta^{(l)}(h^{(l,k-1)})\quad \text{for } k=2,\dots, r_l.
\]
Here, $h^{(l,k)}$ is the intermediate representation after the $k^{th}$ recurrence of layer $l$.

During backpropagation, gradients accumulate across all recurrences of a layer, making the model more sensitive to instabilities such as gradient explosion or vanishing if using a high amount of recurrence steps in a single layer.

Let \(\mathcal{L}\) be the loss function. Define:
\[
J^{(l,k)} \equiv \frac{\partial f_\theta^{(l)}(h^{(l,k-1)})}{\partial h^{(l,k-1)}}, \quad
\delta^{(l,k)} \equiv \frac{\partial \mathcal{L}}{\partial h^{(l,k)}}.
\]

\paragraph{Gradient w.r.t. the Input}
The gradient flowing back to the input \(h^{(l-1)}\) is:
\[
\frac{\partial \mathcal{L}}{\partial h^{(l-1)}} = \left(\prod_{j=1}^{r_l} J^{(l,j)}\right) \delta^{(l,r_l)}.
\]

\paragraph{Gradient w.r.t. the Parameters}
The gradient with respect to the parameters \(\theta^{(l)}\) accumulates contributions from each recurrence:
\[
\frac{\partial \mathcal{L}}{\partial \theta^{(l)}} = \sum_{k=1}^{r_l} \left(\prod_{j=k+1}^{r_l} J^{(l,j)}\right) \delta^{(l,r_l)} \cdot \frac{\partial f_\theta^{(l)}(h^{(l,k-1)})}{\partial \theta^{(l)}}.
\]

\section{Experiments and Results}
\label{sec:experiments}

We use the LLaMA architecture \cite{touvron2023llama} and define a \textbf{reuse map} for per-layer recurrence.\footnote{Code for the modified architecture and training is available at \href{https://github.com/ant-8/Layer-Recurrent-Transformers}{\textcolor{blue}{github.com/ant-8/Layer-Recurrent-Transformers}}}
 Experiments are conducted at two scales: small (1.2M parameters) and large (100M). Our pretraining and test dataset comes from a deduplicated Fineweb-Edu subset \cite{penedo2024finewebdatasetsdecantingweb}.

We train a model for each of the following positional encoding methods: NoPE \cite{kazemnejad2023impact}, RoPE \cite{su2023roformer}, Learned Absolute PE \cite{devlin2019bert}, and ALiBi \cite{press2022trainshorttestlong}. While Learned Absolute PE applies fixed embeddings once, RoPE and ALiBi reapply positional information at every attention step, impacting how recurrence preserves position awareness.

To assess recurrence impact, we test various \textbf{reuse maps} and the evaluate language modeling perplexity (lower is better) on the test set. The small-scale model explores single-layer reuse (e.g., $[2,1,1,1]$, $[1,2,1,1]$, $[1,1,2,1]$, $[1,1,1,2]$) and doubled depth usage (e.g., $[2,2,2,2]$, $[3,2,2,1]$).

\begin{figure}[h]
    \centering
    \includegraphics[width=0.45\linewidth]{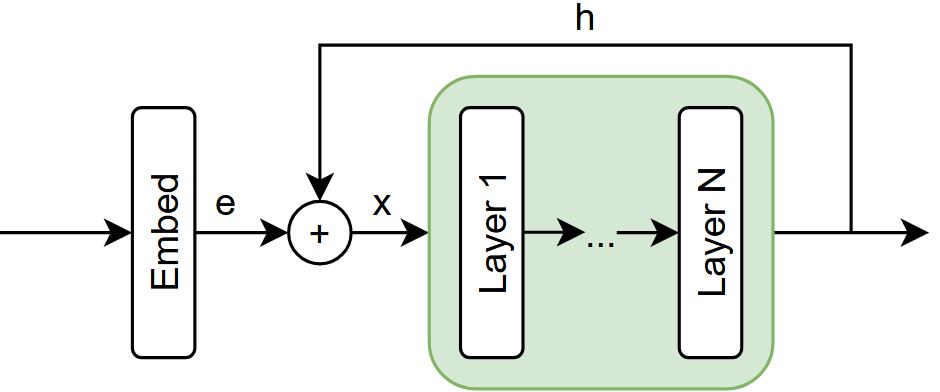}
    \caption{Diagram illustrating block recurrence from small-scale experiments. Unlike ILR, recurrence is applied across the entire stack rather than per layer.}
    \label{fig:block_recurrence}
\end{figure}

We also evaluate \textbf{block recurrence}, where a block of layers is iterated for $r = 2$ steps, with hidden states sequentially propagating through all layers at each step, as seen in related works. A notable difference lies in the state mapping: Geiping et al. ~\cite{geiping2025scaling} employ a learned adapter to map the initial embedding $e$ and the current hidden state $h$ into the input $x$ at each recurrence step. In contrast, implementations in other works~\cite{yang2024looped, fan2024looped} set $x = h + e$, which we adopt in our approach as well. A visual representation of block recurrence is provided in Figure~\ref{fig:block_recurrence}.

For the large-scale model, we train and evaluate a single reuse mapping, guided by results from the small-scale experiments. Due to computational constraints, we selected the optimal configuration that reuses only one layer ($[1,2,1,\dots,1]$).
\newpage
\subsection{Results}
\begin{table}[h]
    \centering
    \renewcommand{\arraystretch}{1.0}
    \begin{tabular}{l|c|c|c|c|c|c}
        \toprule
        \textbf{Recurrence Strategy} & \textbf{Reuse Map} & \textbf{Model Size} & \textbf{NoPE} & \textbf{RoPE} & \textbf{Learned} & \textbf{ALiBi} \\
        \midrule
        Baseline (No recurrence) & -- & Small & 16.57 & 15.56 & 14.98 & 14.38 \\
        Block Recurrence ($r=2$) & -- & Small & 15.29 & 14.12 & 14.27 & 14.23 \\
        ILR & $[2,1,1,1]$ & Small & 15.17 & 14.4 & 14.42 & 13.87 \\
        ILR & $[1,2,1,1]$ & Small & 15.84 & 13.93 & 14.39 & 14.02 \\
        ILR & $[1,1,2,1]$ & Small & 16.54 & 14.3 & 14.81 & 13.92 \\
        ILR & $[1,1,1,2]$ & Small & 16.98 & 15.02 & 14.94 & 14.23 \\
        ILR & $[1,1,2,4]$ & Small & 17.54 & 14.24 & 15.0 & 14.13 \\
        ILR & $[1,2,2,3]$ & Small & 15.59 & 13.96 & 14.25 & 13.88 \\
        ILR & $[2,2,2,2]$ & Small & 15.07 & 14.15 & \textbf{14.17} & 13.76 \\
        ILR & $[3,2,2,1]$ & Small & \textbf{14.62} & 14.57 & 14.31 & 13.64 \\
        ILR & $[4,2,1,1]$ & Small & 14.64 & \textbf{13.77} & 14.2 & \textbf{13.63} \\

        \midrule
        Baseline (No recurrence) & -- & Large & 18.09 & 16.77 & 17.64 & 17.16 \\
        ILR & $[1, 2, 1, ..., 1]$ & Large & \textbf{17.97} & \textbf{16.64} & \textbf{17.54} & \textbf{16.98} \\
        \bottomrule
    \end{tabular}
    \caption{Perplexity results for different reuse maps in small and large-scale models (tested on trained sequence length of 1024). Lower is better (best in bold).}
    \label{tab:perplexity_results}
\end{table}

\begin{table}[h]
    \centering
    \renewcommand{\arraystretch}{1.0}
    \begin{tabular}{l c c}
        \toprule
        \textbf{Recurrence Strategy} & \textbf{Small Model} & \textbf{Large Model} \\
        \midrule
        Baseline (No recurrence) & 4.13 $\times 10^{15}$ & 693.6 $\times 10^{15}$ \\
        Reuse single layer & 5.16 $\times 10^{15}$ & 769.8 $\times 10^{15}$ \\
        Doubled depth & 8.24 $\times 10^{15}$ & -- \\
        \bottomrule
    \end{tabular}
    \caption{Training FLOPs for different configurations.}
    \label{tab:flops_comparison}
\end{table}

\subsection{Observations and Analysis}

Our experiment highlights a key trend regarding the impact of layer reuse on transformer performance.

\textit{Applying recurrence to earlier layers usually yields the largest perplexity gains}, aligning with prior research \cite{kovaleva2019bert, logitlens2020} showing early layers are most influential in encoding core representations while later layers refine them. In our small-scale model, prioritizing early-layer reuse (\([4,2,1,1]\)) reduced perplexity from \textbf{16.57} to \textbf{14.62} (NoPE) and \textbf{14.38} to \textbf{13.63} (ALiBi) as shown in Table \ref{tab:perplexity_results}. Other early-focused configurations, like \([2,1,1,1]\), also improved results, supporting the benefit of reinforcing lower-layer representations.

As seen in Table \ref{tab:flops_comparison}, recurrence increases computational overhead. To improve computational efficiency, future work could investigate \textit{adaptive recurrence mechanisms} that selectively reuse network layers based on input complexity, such as the difficulty of a task specified in a prompt to an instruction-tuned language model.

\section{Limitations}

While layer reuse improves perplexity without increasing parameter count, it introduces challenges that warrant further investigation.

One limitation is the \textit{increased computational cost}. Since reused layers undergo multiple forward passes, training and inference require more computation.

A major challenge in our investigation is the \textit{limited training steps for the large model}, which was constrained by available compute and time resources. The effective training size of 3B tokens (500M tokens $\times$ 6 epochs) may be insufficient for a 100M-parameter model to fully exploit it. Scaling laws \cite{hoffmann2022training} suggest that larger models require more compute to benefit from increasing model sizes.

\section{Conclusion}

We investigated ILR in transformers, where select layers are re-entered within a single forward pass. Our results show that layer reuse improves perplexity without increasing model size, with early layers benefiting the most.

However, reuse increases computational cost, introducing a compute-performance trade-off. While ILR provides a viable method for enhancing transformers in parameter-constrained settings, finding optimal reuse maps remains a challenge, especially for larger models with many layers. Future work should explore strategies for efficiently discovering optimal recurrence distributions across layers, reducing the need for exhaustive experimentation. This is particularly important for scaling ILR to large models, where the number of possible reuse maps grows significantly.

\appendix
\section{Experimental Details}
\label{app:model_config_hyperparam}

We provide tables summarizing the model configurations (Table \ref{tab:model_configs}), training hyperparameters (Table \ref{tab:training_hyperparams}), and dataset usage (Table \ref{tab:training_data}). We train and evaluate two decoder-only LLaMA-based transformers at different scales to assess the impact of intra-layer recurrence across model capacities.

\begin{table}[h]
    \centering
    \renewcommand{\arraystretch}{1.1}
    \begin{tabular}{lcccccc}
        \toprule
        \textbf{Model} & \textbf{Params} & \textbf{Hidden Dim} & \textbf{Layers} & \textbf{Heads} & \textbf{Vocab} \\
        \midrule
        Small & 1.2M & 128 & 4 & 4 & 1,024 \\
        Large & 100M & 768 & 8 & 8 & 32,000 \\
        \bottomrule
    \end{tabular}
    \caption{Model configurations.}
    \label{tab:model_configs}
\end{table}

\begin{table}[h]
    \centering
    \renewcommand{\arraystretch}{1.2}
    \begin{tabular}{l|c|c}
        \toprule
        \textbf{Hyperparameter} & \textbf{Small Model} & \textbf{Large Model} \\
        \midrule
        Optimizer & AdamW & AdamW \\
        Learning Rate & $3 \times 10^{-3}$ & $1 \times 10^{-3}$ \\
        LR Warmup & 10\% of steps & 2\% of steps \\
        Scheduler & Cosine & Cosine \\
        Batch Size & 64 & 64 \\
        Gradient Clip & 1.0 & 1.0 \\
        \bottomrule
    \end{tabular}
    \caption{Training hyperparameters for small and large models.}
    \label{tab:training_hyperparams}
\end{table}

\begin{table}[h]
    \centering
    \renewcommand{\arraystretch}{1.1}
    \begin{tabular}{l|c|c}
        \toprule
        \textbf{Model} & \textbf{Train Tokens} & \textbf{Epochs} \\
        \midrule
        Small & 500M & 1 \\
        Large & 500M & 6 \\
        \bottomrule
    \end{tabular}
    \caption{Dataset usage and training duration.}
    \label{tab:training_data}
\end{table}

% All references should be stored in the file "references.bib".
% That call to use that file is in "cai.cls". 
% Please do not modify anything below this line.

\printbibliography[heading=subbibintoc]

\end{document}